# Optimal Bangla Keyboard Layout using Data Mining Technique

S. M. Kamruzzaman[1]
Md. Hijbul Alam[2]
Abdul Kadar Muhammad Masum[3]
Mohammad Mahadi Hassan[4]

## Abstract

This paper presents an optimal Bangla Keyboard Layout, which distributes the load equally on both hands so that maximizing the ease and minimizing the effort. Bangla alphabet has a large number of letters, for this it is difficult to type faster using Bangla keyboard. Our proposed keyboard will maximize the speed of operator as they can type with both hands parallel. Here we use the association rule of data mining to distribute the Bangla characters in the keyboard. First, we analyze the frequencies of data consisting of monograph, digraph and trigraph, which are derived from data wire-house, and then used association rule of data mining to distribute the Bangla characters in the layout. Experimental results on several data show the effectiveness of the proposed approach with better performance.

## 1 Introduction

The usefulness of computer is increasing rapidly. It is the fastest growing industry in the world. New invention of technologies makes it faster. The billions of instructions can be done within a second nowadays. But the technology of input device like keyboard has not been changed much. The typing speed is very slow in respect of computer's other devices. Its main cause is the complex and insufficient keyboard layout and human limitations [8] [9]. A scientific keyboard layout with equal hands load and maximum hand switching can reduce this problem. The use of Bangla is increasing day by day in everyday life. Specially in data entry and printing sector. But there is no scientific Bangla keyboard layout at present and very few research works have been done in this field. We use the concept of data mining association rule to design Bangla keyboard layout that shows optimal and better performance than the existing keyboard layout.

In this paper we have tried to design a Bangla keyboard layout using the association rule of data mining. We have collected the data from various Bangla documents and we have used the association rule to extract the association of Bangla letters each other. And finally we have designed a Bangla keyboard layout with equal hands load and maximum hand switching.

## 2 Background Study

### 2.1 Data Mining

Data mining refers to extracting or "mining" knowledge from large amounts of data [1]. It can also be named by "knowledge mining form data". Nevertheless, mining is a vivid term characterizing the process that finds a small set of precious nuggets from a great deal of raw material. There are many other terms carrying a similar or slightly different meaning to data mining, such as knowledge mining from databases, knowledge extraction, data/ pattern analysis, data archaeology, and data dredging.

[1] Assistant Professor, Department of Computer Science and Engineering, Manarat International University, Dhaka-1212, Bangladesh
[2] Department of Computer Science and Engineering, International Islamic University Chittagong, Chittagong-4203, Bangladesh
[3] As above
[4] As above

Many people treat data mining as a synonym for another popularly used term, Knowledge Discovery in Databases, or KDD [2]. Alternatively, data mining is also treated simply as an essential step in the process of knowledge discovery in databases.

The fast-growing, tremendous amount of data, collected and stored in large and numerous databases, has far exceeded our human ability for comprehension without powerful tools [4]. In such situation we become data rich but information poor. Consequently. Important decisions are often made based not on the information-rich data stored in databases but rather on a decision maker's intuition, simply because the decision maker does not have the tools to extract the valuable knowledge embedded in the vast amounts of data. In addition, current expert system technologies rely on users or domain experts to manually input knowledge into knowledge bases [5]. Unfortunately, this procedure is prone to biases and errors, and is extremely time consuming and costly. In such situation data mining tools can perform data analysis and may uncover important data patterns, contributing greatly to business strategies, knowledge bases, and scientific or medical research.

**2.2 Association Rule**

Association rule mining finds interesting association or correlation relationships among a large set of data items. In short association rule is based on associated relationships. The discovery of interesting association relationships among huge amounts of transaction records can help in many decision-making processes. Association rules are generated on the basis of two important terms namely minimum support threshold and minimum confidence threshold.

Let us consider the following assumptions to represent the association rule in terms of mathematical representation,
J = {i1, i2, . . . , im} be a set of items. Let D the task relevant data, be a set of database transactions where each transaction T is a set of items such that $T \subseteq J$. Each transaction is associated with an identifier, called TID. Let A be a set of items. A transaction T is said to contain A if and only if $A \subseteq T$. An association rule is an implication of the form $A \Rightarrow B$, where $A \subset J$, $B \subset J$, and $A \cap B = \Phi$. The rule $A \Rightarrow B$ holds in the transaction set D with support s, where s is the percentage of transactions in D that contain $A \cup B$ (i.e. both A and B). This is taken to be the probability, P ($A \cup B$). The rule $A \Rightarrow B$ has confidence c in the transaction set d if c is the percentage of transaction in D containing A that also contain B. This is taken to be the conditional probability, P (B | A). That is,

$$\text{support } (A \Rightarrow B) = P\,(A \cup B) \text{ and confidence } (A \Rightarrow B) = P\,(B \mid A)$$

Association Rules that satisfy both a minimum support threshold and minimum confidence threshold are called strong association rules. A set of items is referred to as an itemset. In data mining research literature, "itemset" is more commonly used than "item set". An itemset that contains k items is a k-itemset. The occurrence frequency of an itemset is the number of transactions that contain the itemset. This is also known, simply as the frequency, support count, or count of the itemset. An itemset satisfies minimum support if the occurrence frequency of the itemset is greater than or equal to the product of minimum support and the total number of transactions in D. The number of transactions required for the itemset to satisfy minimum support is therefore referred to as the minimum support count. If an itemset satisfies minimum support, then it is a frequent itemset. The set of frequent k-itemsets is commonly denoted by Lk.

Association rule mining is a two-step process, which includes:

1. Find all Frequent Itemsets
2. Generate Strong Association Rules from the Frequent Itemsets

**2.3 The Apriori Algorithm**

Apriori is an influential algorithm for mining frequent itemsets for Boolean association rules [1] . The name of the algorithm is based on the fact that the algorithm uses prior knowledge of frequent itemset properties. Association rule mining is a two steps process [7].

i) Find all frequent itemsets: By definition, each of these itemsets will occur at least as frequently as a pre-defined minimum support count.

ii) Generate strong Association rules from the frequent itemsets: By definition, these rules must satisfy minimum confidence.

Apriori employees an iterative approach known as a level-wise search, where *k*-itemsets are used to explore *(k+1)*-itemsets. First, the set of frequent 1-itemsets is found. This set is denoted $L_1$. $L_1$ is used to find $L_2$, the set of frequent 2-itemsets, which is used to find $L_3$, and so on, until no more frequent *k*-itemsets can be found. The finding of each $L_k$ requires one full scan of the database. An important property called Apriori property, based on the observation is that, if an itemset *I* is not frequent, that is, $P(I) < \text{min_sup}$ then if an item *A* is added to the itemset *I*, the resulting itemset (i.e., $I \cup A$) cannot occur more frequently than *I*. Therefore, $I \cup A$ is not frequent either, that is, $P(I \cup A) < min_sup$. To understand how Apriori property is used in the algorithm, let us look at how $L_{k-1}$ is used to find $L_k$. A two-step process is followed, consisting of join and prune actions.

The join step: To find $L_k$, a set of candidate *k*-itemsets is generated by joining $L_{k-1}$ with itself. This set of candidates is denoted by $C_k$. Let $l_1$ and $l_2$ be itemsets in $L_{k-1}$ then $l_1$ and $l_2$ are joinable if their first *(k-2)* items are in common, i.e., $(l_1[1]=l_2[1]) . (l_1[2]=l_2[2]) \ldots\ldots (l_1[k-2]=l_2[k-2]) . (l_1[k-1]<l_2[k-1])$.

The prune step: $C_k$ is the superset of $L_k$. A scan of the database to determine the count of each candidate in $C_k$ would result in the determination of $L_k$ (itemsets having a count no less than minimum support in $C_k$). But this scan and computation can be reduced by applying the Apriori property. Any *(k-1)*-itemset that is not frequent cannot be a subset of a frequent *k*-itemset. Hence if any *(k-1)*-subset of a candidate *k*-itemset is not in $L_{k-1}$, then the candidate cannot be frequent either and so can be removed from $C_k$..

**2.4 Illustration of Apriori Algorithm**

Consider an example of Apriori, based on the following transaction database, D of Figure: 1, with 9 transactions, to illustrate Apriori algorithm. In the first iteration of the algorithm, each item is a member of the set of candidate 1-itemsets, *C1* The algorithm simply scans all of the transactions in order to count the number of occurrences of each item. If minimum support count is set to 2,frequent 1-itemsets, *L1 ,* can then be determined from candidate 1-itemsets satisfying minimum support [6].

To discover the set of frequent 2-itemsets, *L2, the* algorithm uses *L1* | *L1* to generate a candidate set of 2-itemsets (Figure: 4).

The transactions D are scanned and the support count of each candidate itemset in *C2* is accumulated (Figure: 5).

The set of 2-itemsets, *L2* (Figure: 6), is then determined, consisting of those candidate 2-itemsets in *C2* having minimum support.

The generation of the set of candidate 3-itemsets, *C3* , is detailed in Figure: 7 to *Itemset Sup. count* Here *C3 = L2* | *L2* = {{I1,I2,I3},{I1,I2,I5}, {I1,I3,I5}, {I2,I3,I5}, {I2,I4,I5}}.Based on the Apriori property that all subsets of a frequent itemset must also be frequent, the resultant candidate itemsets will be as in Figure: 7.

The transaction in D are scanned in order to determine *L3* , consisting of those candidate 3-itemsets in *C3* having minimum support (Figure: 9).

The algorithm uses *L3* | *L3* to generate a candidate set of 4-itemsets, *C4* . Although the join results in {{I1,I2,I3,I5}}, this itemset is pruned since its subset {{I2,I3,I5}} is not frequent. Thus, *C4* = {}, and the algorithm terminates.

Let us consider an example of Apriori, based on the following transaction database, D of figure 4.1, with 9 transactions, to illustrate Apriori algorithm.

| TID | List of item_IDs |
|---|---|
| T100 | I1, I2, I5 |
| T200 | I2, I4 |
| T300 | I2, I3 |
| T400 | I1, I2, I4 |
| T500 | I1, I3 |
| T600 | I2, I3 |
| T700 | I1, I3 |
| T800 | I1, I2, I3, I5 |
| T900 | I1, I2 , I3 |

**Figure 1.** Transactional Data

| Itemset | Sup. count |
|---|---|
| {I1} | 6 |
| {I2} | 7 |
| {I3} | 6 |
| {I4} | 2 |
| {I5} | 2 |

**Figure 2.** Generation of C1

| Itemset | Sup. count |
|---|---|
| {I1} | 6 |
| {I2} | 7 |
| {I3} | 6 |
| {I4} | 2 |
| {I5} | 2 |

**Figure 3.** Generation of L1

In the first iteration of the algorithm, each item is a number of the set of candidate 1-itemsets, C1. The algorithm simply scans all of the transactions in order to count the number of occurrences of each item.

Suppose that the minimum transaction support count required is 2 (i.e.; min_sup = 2/9 = 22%). The set of frequent 1-itemsets, L1, can then be determined. It consists of the candidate 1-itemsets satisfying minimum support.
To discover the set of frequent 2-itemsets, L2, the algorithm uses L1 | L2 to generate a candidate set of 2-itemsets, C2.
The transactions in D are scanned and the support count of each candidate itemset in C2 is accumulated.

| Itemset | Sup.count |
|---|---|
| {I1, I2} | 4 |
| {I1, I3} | 4 |
| {I1, I4} | 1 |
| {I1, I5} | 2 |
| {I2, I3} | 4 |
| {I2, I4} | 2 |
| {I2, I5} | 2 |
| {I3, I4} | 0 |
| {I3, I5} | 1 |
| {I4, I5} | 0 |

**Figure 4.** Generation of C2

| Itemset | Sup.count |
|---|---|
| {I1, I2} | 4 |
| {I1, I3} | 4 |
| {I1, I5} | 2 |
| {I2, I3} | 4 |
| {I2, I4} | 2 |
| {I2, I5} | 2 |

**Figure 5.** Generation of L2

| **Itemset** | **Sup.count** |
|---|---|
| {I1, I2, I3} | 2 |
| {I1, I2, I5} | 2 |

**Figure 6.** Generation of C3

| **Itemset** | **Sup.count** |
|---|---|
| {I1, I2, I3} | 2 |
| {I1, I2, I5} | 2 |

**Figure 7.** Generation of L3

The set of frequent 2-itemsets, L2, is then determined, consisting of those candidate-itemsets in C2 having minimum support.
The generation of the set of candidate 3-itemsets, C3 is observed in Fig 4.6 to Fig 4.7. Here C3 = L1 | L2 = {{l1, l2, l3}, {l1, l2, l5}, {l1, l3, l5}, {l2, l3, l5}, {l2, l4, l5}}. Based on the Apriori property that all subsets of a frequent itemset must also be frequent, we can determine that the four latter candidates cannot possibly be frequent.
The transactions in D are scanned in order to determine L3, consisting of those candidate 3-itemsets in C3 having minimum support.

The algorithm uses L3 | L3 to generate a candidate set of 4-itemsets, C4. Although the join results in {{l1, l2, l3, l5}}, this itemset is pruned since its subset {{l2, l3, l5}} is not frequent. Thus, C4 = {}, and the algorithm terminates.

## 3 Proposed Method

At first we calculate the monograph, digraph from the frequencies of characters from the Bangla documents. And then we find the association of a character with the determined left hand typed letter sets and right hand typed letter sets according to association rule and distribute it in opposite side for better or maximize hand switching as described in the algorithm at section 4. Finally, we distribute the most frequent word closer to finger for easy and first access. Before stating the algorithm we have to explain some keyword in brief. The meaning of support and confidence is stated earlier in section 2. We introduce four new terms for our purpose. Left support of a letter means the cumulative support of the letter to the set of left sets letter. Left confidence of a letter means the cumulative confidence of the letter to the right sets letter. Similarly right support and confidence carry the meaning as well.

### 3.1 Proposed Algorithm

1. Find out the most frequent itemset of monograph (letter) in descending order
2. Distribute 1st and 4th letter in the right hand (as 1st letter found twice of second)
3. Distribute 2nd and 3rd letter in the left hand
4. Now take the letter one by one starting from 5th letter and distribute it in left hand or right depend on the following criteria
    a. Find the cumulative support and confidence of this character with all right hand letter distributed till now and labeled it as right support and right confidence
    b. Similarly find left support and left confidence for the same letter
    c. if left support > right support and left confidence > right confidence then distribute the letter in right side otherwise in left side.
5. Repeat the step 4 until all letters are distributed.

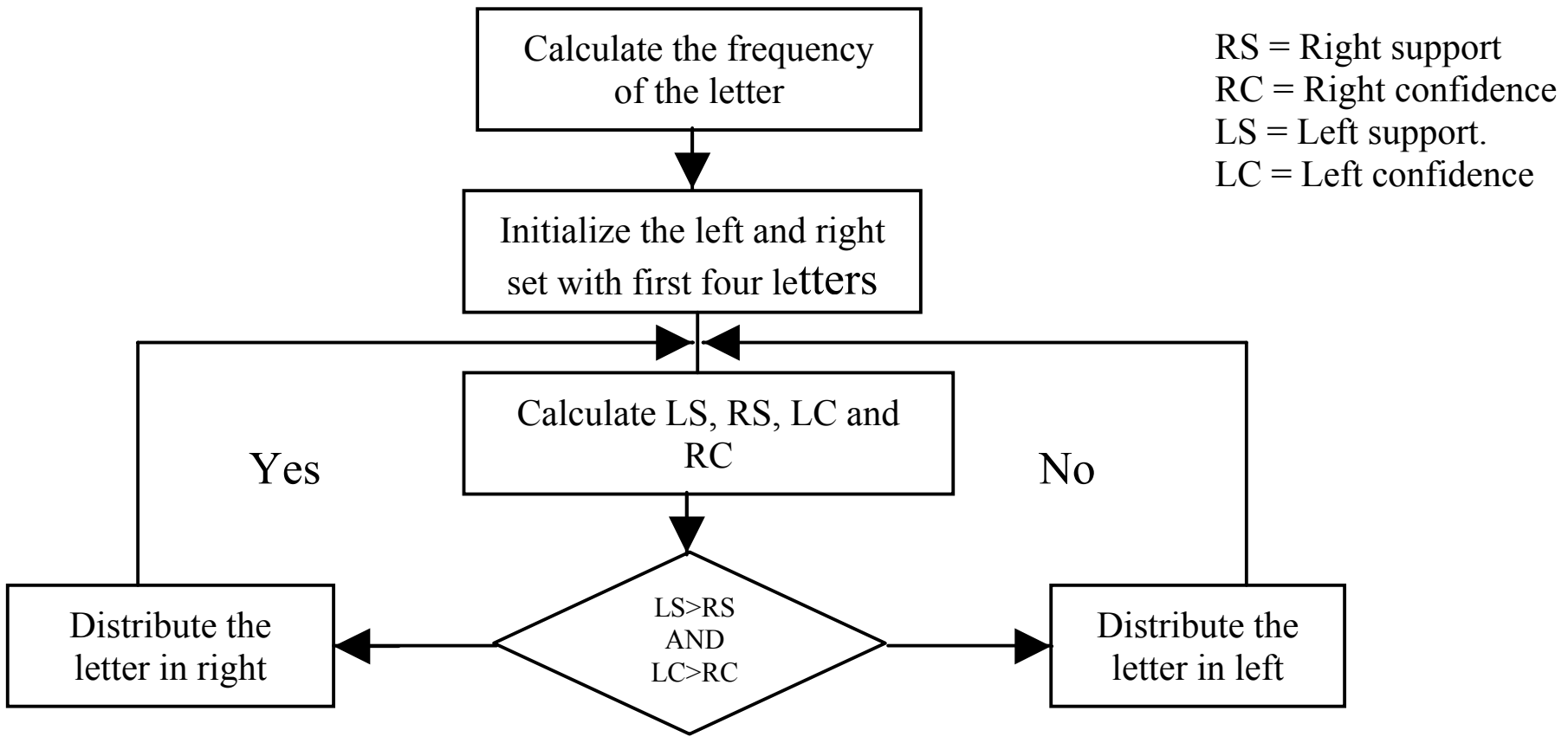

**Figure 8.** Flow Chart of the proposed algorithm

## 4 Experimental Result

About more than 8 lakh letters are extracted from various types of documents such as prose, text book, novel, religious book etc. From this data we find out monographs, digraphs and then determine support and confidence to mine the association.

Table 1: The first ten monograms of our experiment

| Letter | Frequency | Percentage |
|---|---|---|
| া | 74300 | 9.039875 |
| ে | 45525 | 5.538901 |
| র | 41844 | 5.091044 |
| ি | 37010 | 4.502904 |
| ক | 31214 | 3.797721 |
| ই | 28996 | 3.527863 |
| ব | 28212 | 3.432476 |
| ত | 21451 | 2.609884 |
| প | 18419 | 2.240989 |
| ম | 17202 | 2.092920 |

So according to the 1-3 steps of the algorithm we can initialize the left and right side characters, which are as follows: Right { া, ি }, left { ে, র } Now we have to take the decision in which set the 5th character will lies. The decision is depend on the step 4.

Table 2: The association of ka( ক ) with some letters are as follows

| Digraph | Frequency | Support | Confidence |
|---|---|---|---|
| কে | 8316 | 1.011785 | 21.717897 |
| কা | 8000 | 0.973338 | 20.892638 |
| কর | 4134 | 0.502972 | 10.796271 |
| কি | 3094 | 0.376438 | 8.080228 |
| এক | 2062 | 0.250878 | 5.385077 |
| তক | 1231 | 0.149772 | 3.214855 |
| কশ | 1153 | 0.140282 | 3.011151 |

The support and confidence of K with left item set { ে, র } is

| | | |
|---|---|---|
| কে | 1.011785 | 21.717897 |
| কর | 0.502972 | 10.796271 |
| (After addition) | 1.514757 | 32.514168 |

So left support (cumulative) is 1.5154757
left confidence (cumulative) is 32.514168

The support and confidence of K with right item set { া, ি } is

| | | |
|---|---|---|
| কা | 0.973338 | 20.892638 |
| কি | 0.376438 | 8.080228 |
| (After addition) | 1.349776 | 28.972866 |

So right support (cumulative) is 1.349776

Right confidence is 28.972866. In this example left support is greater then right support and left confidence is greater right confidence. Thus ক will go to right item sets. As a result the members of itemsets are increased. The sets Right { া, ি ক } and left { ে, র } Thus all letter are distribute in left and right side according to the descending order of the frequency of the letter. After executing the program we got the following two sets

Right{ া ি ক স ল য় ম শ ী ্ এ থ উ ণ হ ঃ ড় ক্ ঠ ৈ ঢ় ঐ ঔ ঞ ঊ ঔ }
Left { ে র ন ব ত প ম হ অ । য জ গ ধ ট ভ ং চ ও ই খ ৎ ৃ ঘ ঁ ঝ ড ী ঈ ষ ঢ ‹ }

Then we distribute the letter according to the most frequent occurring letter in middle row that is closer to the hand as described in the algorithm at step 4. Our designed Bangla keyboard layout is below.

। অ প ম হ    য় ল স দ শ

ত ব ে র ন    ্ ি া ক

ট থ য জ গ    ৗ ঢ়

with shift key

ঁ ষ থ ৎ ৃ    ঃ ছ ণ ড় ফ

ই ও ভ ং চ    ধ এ ্ উ

ঝ ঈ ঋ ড ী    ঠ ৈ ্ ‹

With ctlr key ঢ ঢ় ঐ ঔ ঞ ঊ ঔ

A Sample of proposed keyboard layout

| “ | 1 | 2 | 3 | 4 | 5 | 6 | 7 | 8 | 9 | 0 |
|---|---|---|---|---|---|---|---|---|---|---|
| Tab | ।<br>Q<br>ৃ | অ<br>W<br>ষ | প<br>E<br>থ | ম<br>R<br>ৎ | হ<br>T<br>ঁ | য়<br>Y<br>ঃ | ল<br>U<br>ছ | স<br>I<br>ণ | দ<br>O<br>ড় | ক<br>P<br>ফ |
| Cap | ত<br>A<br>ই | ব<br>S<br>ও | ে<br>D<br>ভ | র<br>F<br>ং | ন<br>G<br>চ | ্<br>H<br>ধ | ি<br>J<br>এ | া<br>K<br>্ | ক<br>L<br>উ | :<br>; |
| Shift | ট<br>Z<br>ঝ | অ<br>X<br>ঈ | য<br>C<br>ঋ | জ<br>V<br>ড | গ<br>B<br>ী | ঁ<br>N<br>ঠ | ৗ<br>M<br>ৈ | ঢ়<br>্ | ,<br>‹ | . |
| Ctrl | | Alt | Space | | | | Alt | | Ctrl | |

**Figure9.** Proposed keyboard layout

## 5 Comparative Study

Here we show the comparative study and experimental result with our proposed keyboard layout to Bijoy and proposed lay out3 [3] [10]. The layouts are as follows

**Bijoy keyboard layout:**

ঙ য ড প ট  চ জ হ গ ড়

ু ি া ্  ব ক ত দ ;

্র ও ে র ন  স ম , . / ঃ

with shift key

ং য় ঢ ফ ঠ  ছ ঝ ঞ ঘ ঢ়

ৃ ী অ ।  ভ খ থ ধ :

্য ী ৈ ল ণ  ষ শ < > ? ৎ

**Proposed key board layout 3:**

ছ অ দ ী য  ল স য় ু এ

ে র ক ত ম  া ন ি ব প

ছ শ ্র জ খ  ই । গ ষ

With shift key

ফ ড় ঈ ং ঝ  ঘ ঠ

ট ভ ধ ৃ ং ও  থ চ উ ণ

ড ৈ ঢ় ঙ ঊ  ?: ঔ ৈ ঐ,

Table 3: comparative study with other keyboard

| Name | Hand switching | left hand load | Right hand load | Not determine |
|---|---|---|---|---|
| Proposed Optimal keybord layout | 410113 | 380058 | 340903 | 133290 |
| Bijoy keyboard layout | 358873 | 475556 | 242526 | 138643 |
| Proposed layout 3 | 358672 | 319946 | 363077 | 173702 |

From the above table we can easily find out that our keyboard layout is optimal. Here we work with total 856725 character (with out space). In Bijoy the load in left is very large than right hand. And hand switching is also smaller than us. Increasing of the data this ratio also increased highly. On the other our keyboard layout is optimal because we design the layout by mining the proper association of letter each other in the various kinds of thoroughly. Not only depend on the frequently occurring monograph, digraph, etc. As for why our keyboard layout shows maximum hand switching than also proposed keyboard layout 3. In the above the table not determining characters are large because here we do not consider 0 to 10, united font etc.From the above data it is clear that our proposed layout divides the load on both hands equally and hand switching is also maximize. Comparing with other keyboard layout we can say that, our proposed layout is optimal.

## 6 Conclusion

The use of Bangla keyboard layout is increasing day by day. So it is the time to research on this important field. The Bangla academy also can come forward to finalize such an important matter and would produce an optimal keyboard layout and circular throughout Bangladesh for new generation. Our keyboard layout shows maximum hand switching than any other exiting keyboard and also cost effective because frequently occurred letter is distribute closer to finger. Experimental results show that our proposed layout is more scientific and user friendly. If our keyboard layout is implemented, we hope that people will be benefited.